\pdfoutput=1

\documentclass[11pt]{article}

\usepackage{amsmath,amsfonts,bm}

\def\eqref#1{equation~\ref{#1}}

\def\1{\bm{1}}

\def\vv{{\bm{v}}}

\def\vx{{\bm{x}}}
\def\vy{{\bm{y}}}
\def\vz{{\bm{z}}}

\DeclareMathAlphabet{\mathsfit}{\encodingdefault}{\sfdefault}{m}{sl}
\SetMathAlphabet{\mathsfit}{bold}{\encodingdefault}{\sfdefault}{bx}{n}

\def\sD{{\mathbb{D}}}

\usepackage[]{ACL2023}

\usepackage{times}
\usepackage{latexsym}
\usepackage{graphicx}
\usepackage{subfigure}
\usepackage{booktabs}
\usepackage{float}
\usepackage{cleveref}
\usepackage{algorithm}
\usepackage{algpseudocode}
\usepackage{amsmath,amsfonts,amssymb}
\usepackage{multirow}
\usepackage[T1]{fontenc}

\usepackage[utf8]{inputenc}

\usepackage{microtype}

\usepackage{inconsolata}
\usepackage{ifthen}

\title{Tailoring Instructions to Student's Learning Levels\\ Boosts Knowledge Distillation}

\author{Yuxin Ren$^{1, \dagger,}$\thanks{ \ \ Equal contribution. $^\dagger$ Work done while being at Amazon.} \quad
  Zihan Zhong$^{1, \dagger, \ast}$ \quad 
  Xingjian Shi$^{2, \dagger}$ \quad
  Yi Zhu$^{2, \dagger}$ \quad
  Chun Yuan$^{1}$ \quad
  Mu Li$^{2, \dagger}$ \\
  {$^{1}$Tsinghua University, $^{2}$Boson AI} \\
  \texttt{\{ryx20,zhongzh22\}@mails.tsinghua.edu.cn}, \texttt{\{xingjian,yi,mu\}@boson.ai} \\
\texttt{yuanc@sz.tsinghua.edu.cn} 
}
\begin{document}

\maketitle

\begin{abstract}
It has been commonly observed that a teacher model with superior performance does not necessarily result in a stronger student, highlighting a discrepancy between current teacher training practices and effective knowledge transfer. 
In order to enhance the guidance of the teacher training process, we introduce the concept of distillation influence to determine the impact of distillation from each training sample on the student's generalization ability. 
In this paper, we propose \textbf{L}earning \textbf{G}ood \textbf{T}eacher \textbf{M}atters (LGTM), an efficient training technique for incorporating distillation influence into the teacher's learning process. 
By prioritizing samples that are likely to enhance the student's generalization ability, our LGTM outperforms 10 common knowledge distillation baselines on 6 text classification tasks in the GLUE benchmark. \footnote{Our code is public available at \url{https://github.com/twinkle0331/LGTM}}

\end{abstract}

\section{Introduction}

The recent success of natural language processing (NLP) is driven by the adoption of large-scale pre-trained language models~\citep{devlin2018bert, liu2019roberta, dai2019transformer, yang2019xlnet}. As these models are scaling up in depth and width, they become increasingly computational and storage intensive, making deployment difficult. To address this issue, different methods have been proposed for crafting efficient models with minimal loss in performance, such as weight pruning~\citep{fan2019reducing, li2021differentiable}, network quantization~\citep{kim2021bert, zhang2020ternarybert}, and knowledge distillation (KD)~\citep{sun2019patient, tang2019distilling, sun2020mobilebert}. Among these methods, KD has proven to be effective in various NLP applications~\citep{jiao2020tinybert} and is widely adopted. The idea of KD involves asking a lightweight student model to mimic the output of a large teacher model so as to transfer the knowledge.

Ideally, a teacher with better performance should be able to transfer more knowledge to the student. Therefore in most knowledge distillation algorithms, the teacher network is trained to maximize its own performance. However, multiple studies~\citep{wang2022efficient, cho2019efficacy} have observed that a teacher with higher performance does not necessarily lead to a better-performing student, and may even cause a performance degradation.
\citet{stanton2021does} has attributed this inefficiency in knowledge distillation to challenges during optimization.
As the model capacity gap between the student and the teacher increases, the optimization process becomes more likely to be trapped in local optima~\citep{cho2019efficacy, mirzadeh2020improved}.

One way to address the performance degradation in KD is to update the teacher via feedback from student's performance, also known as learning to teach (L2T)~\citep{fan2018learning, zhou2022bert}. L2T allows the teacher model to adjust its ``teaching agenda'' by interacting with the student. Among the L2T algorithms, online distillation~\citep{zhang2018deep,zhu2018knowledge,shi2020learning} trains the student and teacher concurrently and enforces similarity between their outputs on the training set.  However, online distillation focuses on transferring the knowledge of the teacher to the student on training set without explicitly considering how well the student will perform on validation set. 
On the other hand, meta distillation ~\citep{zhou2022bert,pham2021meta} takes the generalization ability of student on the held-out validation set into account, and guides the teacher's learning process to maximize the generalization ability. However, the optimization objective of meta distillation may result in a degraded teacher model, as it only receives supervision from the student model.

It is well-known that humans are more efficient learners when their teachers provide guidance on the level of attention they should devote to certain problems based on their current knowledge. Similarly, it is possible that a student model could be trained more effectively if it receives such guidance from a teacher.
To accomplish this goal, the teacher should prioritize samples that are likely to enhance the student's generalization ability during training, thus allowing the student to perform better on the held-out validation set.

In this work, inspired by the concept of influence function~\citep{pruthi2020estimating,koh2017understanding}, we propose \textit{distillation influence} to estimate how distilling on each training sample impacts the student's performance on the validation set.
In addition, we are able to interpret existing L2T methods from the perspective of influence function, so as to gain a deeper understanding of their limitations. 
The optimization process of existing L2T methods are often impacted by outliers, because they assign all training samples in the mini-batch the same weight.
Hence, we propose our L2T framework, \textbf{L}earning \textbf{G}ood \textbf{T}eacher \textbf{M}atters (LGTM), which assigns loss weights of the training samples based on their distillation influence.

Extensive experiments have shown that LGTM enables more effective knowledge transfer.

In summary, our contributions are as follows:
\begin{enumerate}
    \item We propose distillation influence to quantify how distilling from each training sample impacts the student's generalization ability.

    \item We introduce finite difference approximation to efficiently incorporate distillation influence into the teacher's learning process.

    \item Comparing to 10 common KD baselines, our proposed LGTM demonstrates consistently better performance on 6 text classification tasks in GLUE benchmark.
\end{enumerate}

\section{Notations}

Suppose we have a teacher model denoted as $T(\cdot; \theta_t)$ and a student model denoted as $S(\cdot; \theta_s)$. 
The corresponding model parameters are $\theta_{t}$ and $\theta_{s}$.
$\eta_{t}$ and $\eta_{s}$ are the learning rates adopted for model update. 
We use $|t|$ and $|s|$ to denote the dimensions of $\theta_{t}$ and $\theta_{s}$, i.e., $\theta_{t}\in \mathbb{R}^{|t|\times1}$ and $\theta_{s}\in \mathbb{R}^{|s|\times1}$.
The time step before and after model parameter updates are denoted as $m$ and $m+1$, respectively. 
It is used to track the evolution of the model parameters during the training process.

Given a labeled training dataset $\sD_{\text{train}}$, a batch of $B^r$ training samples and their corresponding labels are referred to as $\vz^{r}=(\vx^{r},\vy^{r})$, where $r$ indicates training. 
We index each sample in the training batch $\vz^r$ as $\vz^r_{i}$.
Similarly for validation dataset $\sD_{\text{val}}$, we define the batch of  samples as $\vz^e=(\vx^e,\vy^e)$, where $e$ indicates validation.

In addition, we introduce the notation of the Jacobian matrix in the context of working with the chain rule and gradient. 
In particular, let $f: \mathbb{R}^k \rightarrow \mathbb{R}^n$ be a differentiable function, and let $\vv \in \mathbb{R}^k$ be a vector. We use the notation $\frac{\partial f}{\partial \vv} \in \mathbb{R}^{k \times n}$ to represent the Jacobian matrix of $f$, which has dimensions $k \times n$. For simplicity, we annotate $\frac{\partial f}{\partial \vv}$ as $\nabla_{\vv}$.
We use $X^\intercal$ to denote the transpose of the matrix $X$.

\begin{figure}
    \small
    \centering
    \includegraphics[width=\linewidth]{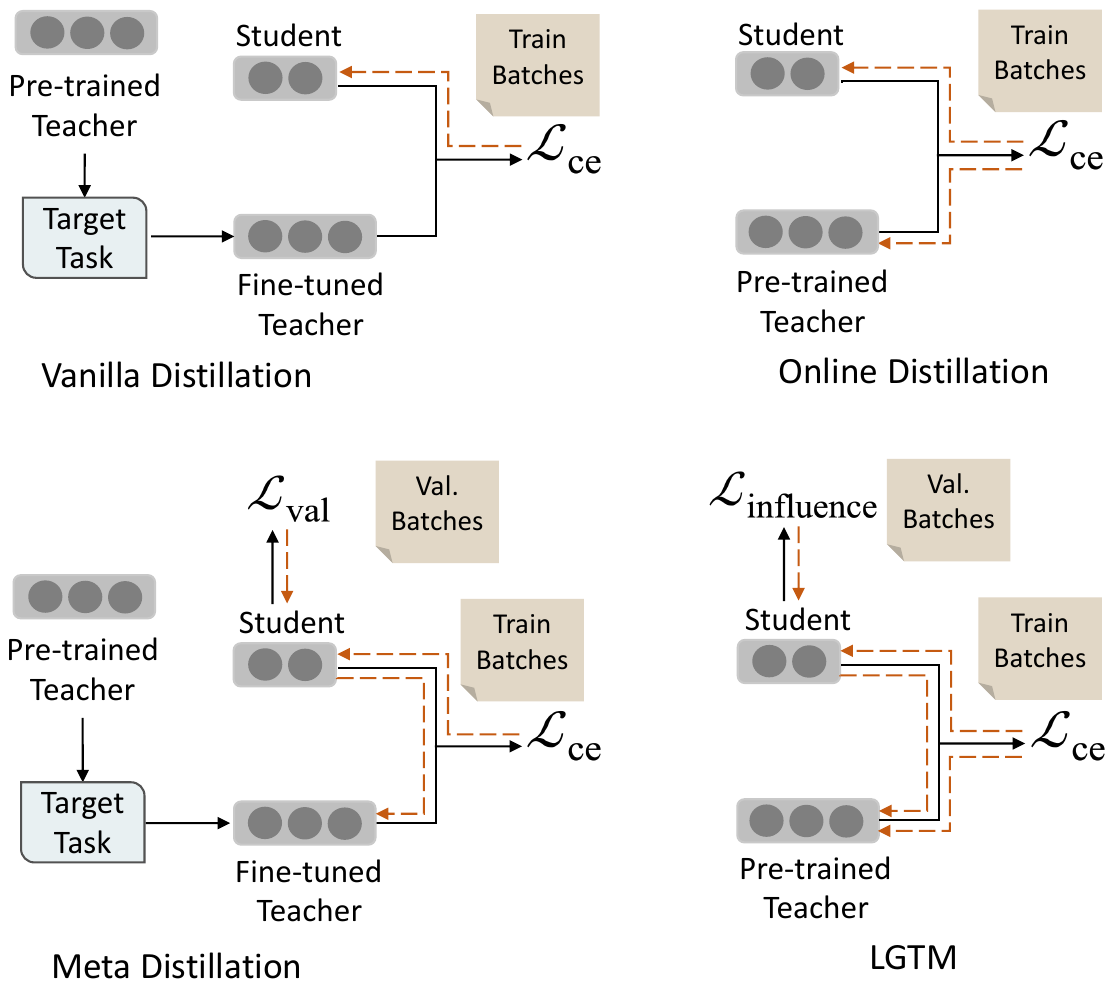}
    \caption{Comparison of vanilla distillation, online distillation, meta distillation and our proposed LGTM. The dotted orange lines show the direction of the gradient flow for model update. Note that vanilla distillation and meta distillation employ a two-stage training pipeline by first fine-tuning the teacher on the target task. Online distillation and LGTM employ a one-stage joint training strategy for both teacher and student.}
    \label{fig:comparision}
\end{figure}

\section{Revisiting Learning to Teach}
In this paper, we focus on task-specific distillation given pre-trained language models. 
Under this setting, the teacher model is already pre-trained in an unsupervised manner and the student model is either derived from part of the teacher model or pre-trained in an unsupervised manner as well. 

\paragraph{Vanilla distillation} The typical approach to knowledge distillation is a two-stage process. It involves first fine-tuning a pre-trained teacher model to maximize its performance on a specific task.
Once the teacher model has converged, a student model is trained to closely imitate the output of the teacher model on the training data. The optimization objective for the student model at each mini-batch is:
\begin{equation}
\begin{aligned}
\mathcal{L}_{\text{s}}(\theta_{s}, \theta_{t},\vz^{r}) = \alpha\mathcal{L}_{\text{ce}}(\vy^r,S(\vx^r;\theta_{s}))\\+(1-\alpha)\mathcal{L}_{\text{ce}}(T(\vx^r;\theta_{t}),S(\vx^r;\theta_{s})).
\end{aligned}
\label{equ:student_loss}
\end{equation}
The update of the student follows:
\begin{equation}
\begin{aligned}
 \theta_{s}^{m+1} = \theta_{s}^{m}-\eta_{s}\nabla_{\theta_{s}} \mathcal{L}_{\text{s}}(\theta_{s}^m, \theta_{t}^m,\vz^r).
\end{aligned}
\end{equation}

The limitation of vanilla distillation is that it does not allow teacher to adjust its behavior according to student's feedback, as the teacher's parameters are fixed during the distillation process.

\paragraph{Online distillation} 

To achieve student-aware distillation, online distillation~\citep{zhang2018deep,zhu2018knowledge, shi2020learning} is proposed which involves the simultaneous fine-tuning of both the student and teacher models in one-stage.

In addition to minimizing the cross-entropy loss with respect to the ground truth labels, the target distribution of the teacher model is constrained to be close to that of the student model through the minimization of the cross-entropy loss between the outputs of the teacher and student models:
\begin{equation}
\begin{aligned}
\mathcal{L}_{\text{t}}(\theta_{t},  \theta_{s},\vz^r) = \alpha\mathcal{L}_{\text{ce}}(\vy^r,T(\vx^r;\theta_{t}))\\+(1-\alpha)\mathcal{L}_{\text{ce}}(T(\vx^r;\theta_{t}),S(\vx^r;\theta_{s})).
\end{aligned}
\label{eqn:teacher}
\end{equation}
The training process involves iteratively updating the parameters of both models:
\begin{equation}
\begin{aligned}
\theta_{t}^{m+1} &= \theta_{t}^m - \eta_{t} \nabla_{\theta_{t}} \mathcal{L}_{t}(\theta_{t}^m, \theta_{s}^m,\vz^r) \\
 \theta_{s}^{m+1} &= \theta_{s}^m - \eta_{s} \nabla_{\theta_{s}} \mathcal{L}_s(\theta_{s}^m, \theta_{t}^{m+1},\vz^r).
\end{aligned}
\end{equation}
Through iterative update, the student model is able to learn from the learning curve of the teacher model~\citep{shi2020learning}, which improves its performance on the given task.

However, online distillation focuses on transferring the knowledge of the teacher to the student on training set without explicitly considering how well the student model will perform on unseen test data. 
This might lead to the student model only memorizing the training examples without generalizing well to new ones~\citep{zhou2022bert}.

\paragraph{Meta distillation}

Meta distillation~\citep{zhou2022bert,pham2021meta} is a technique that takes into account the feedback from the student model and guides the optimization of the teacher model to maximize the generalization ability of the student. The generalization error of the student model is measured by the cross-entropy loss computed between the ground truth labels and the predictions of the student model on the validation set:
\begin{equation}
\begin{aligned}
\mathcal{L}_{\text{val}}(\theta_s,\vz^e)=\mathcal{L}_{\text{ce}}(\vy^e,S(\vx^e,\theta_s)).
\end{aligned}
\end{equation}

Meta distillation decomposes models' learning process into two stages.
The first stage is to fine-tune a good teacher on task-specific data similar to vanilla distillation, while the second stage involves iterative update of the teacher and student models. 
Note that compared to online distillation, meta distillation obtains the student feedback from validation data, not training data.

During the second stage, the student model is first updated through the standard distillation process by minimizing the distillation loss in \cref{equ:student_loss}. 
Then the teacher model is optimized to minimize the updated student's loss on the held-out validation set, which ensures it is able to guide the student towards better generalization. 
During this process, the teacher is only trained for the purpose of knowledge transfer.
Formally, the student model is updated as follows:

\begin{equation}
\begin{aligned}
\theta_{s}^{m+1} = \theta_{s}^{m}-\eta_{s}\nabla_{\theta_{s}} \mathcal{L}_{\text{s}}(\theta_{s}^m, \theta_{t}^m,\vz^r).
\end{aligned}
\end{equation}

The teacher model is then updated as follows:
\begin{equation}
 \theta_{t}^{m+1}=\theta_{t}^{m}-\eta_{t}\nabla_{\theta_{t}} \mathcal{L}_{\text{val}}(\theta_{s}^{m+1},\vz^e),
\end{equation}

However, the optimization objective of meta distillation can result in a degraded teacher model because it only receives supervision from the student. 
This will prevent the teacher model from continuing to learn and improve in the second stage, thus impeding its ability to adapt to new data.

\section{Methods}
\label{sec:methods}

To overcome the aforementioned limitations, we introduce our  L2T framework, \textbf{L}earning \textbf{G}ood \textbf{T}eacher \textbf{M}atters (LGTM) to enable more effective knowledge distillation. We first introduce \emph{distillation influence}, which estimates how much will the student's  performance on validation data change if we put one training sample in the knowledge distillation process.

Afterwards, we introduce an efficient training method based on finite difference approximation for incorporating distillation influence into the teacher's update. 
Finally, we interpret current L2T methods from the perspective of influence function.

\paragraph{Distillation influence} Influence function~\citep{pruthi2020estimating,koh2017understanding} is a way of measuring the influence of training samples on the model's predictions.  It can be utilized to identify instances that have a disproportionate effect on the model's behavior, whether due to their status as outliers or due to incorrect labeling~\citep{jia2019towards,ghorbani2019data,hara2019data}. By calculating the influence function for a particular example, it is possible to estimate the extent to which the model's prediction would be altered as a result of operations on that sample. 

In vanilla distillation, for the student model, we derive the distillation influence of $\vz_i^r$ as the gradient similarity between the training sample $\vz_i^r$ and the validation batch $\vz^e$: 
\begin{equation}
\small
\begin{aligned}
 \mathcal{I}_{\text{distill}}(\vz^r_i,\vz^e) =&
 \nabla_{\theta_{s}}\mathcal{L}_{\text{ce}}(T(\vx^r_i;\theta_{t}^m),S(\vx^r_i;\theta_{s}^m))^\intercal \\ 
 &\nabla_{\theta_{s}}\mathcal{L}_{\text{ce}}(\vy^{e}, S (\vx^{e}; \theta_s^{m+1}))
\end{aligned}
\end{equation}
The detailed derivation can be found in \cref{sec:influence}. The influence reflects how well the knowledge gained from a particular sample generalizes. It follows that the teacher should focus on teaching the student to capture training samples that have the highest distillation influences.

In order to incorporate the per-sample influence into knowledge distillation, we adjust the loss weight of each sample based on its distillation influence. This allows us to determine the relative importance of each sample, and helps to control how much each sample contributes to the teacher's learning process.
Samples that are deemed to be more beneficial for the student's generalization are assigned higher weights. 
Then we propose training the teacher using the following objective:
\begin{equation}
\small
\begin{aligned}
 \mathcal{L}_{\text{influence}} = \frac{1}{B^r} \sum_{i=1}^{B^{r}} w_i {\mathcal{L}_{\text{ce}} ((T(\vx_i^r;\theta_{t}^m),S(\vx_i^r;\theta_{s}^m))},
\end{aligned}
\end{equation}
where $w_i=\mathcal{I}_{\text{distill}}(\vz_i^r,\vz^e)$.
By including the influence in the knowledge distillation loss function, we can tailor the training process to better suit the characteristics of the target task.

\begin{algorithm}[!t]
\small
\caption{LGTM}\label{alg:flow}
\begin{algorithmic}[1]
\Require student $\theta_s$, teacher $\theta_t$, training set $\sD_{\text{train}}$, validation set $\sD_{\text{val}}$
\Require $\eta_s$, $\eta_t$: learning rate for the student and the teacher %
\Require $\epsilon$: a small scalar%
\Require $M$: the maximum number of the training steps%

\While{$step < M$}
\State Sample a batch of training set $\vz^r = (\vx^r, \vy^r) \sim \sD_{\text{train}}$
\State Copy student parameter $\theta_s$ to student $\theta_s'$
\State Update $\theta_s'$: $\theta_s' \gets \theta_s - \eta_s \nabla_{\theta_s'}  \mathcal{L}_s(\theta_s', \theta_t,\vz^r) $
\State Sample a batch of validation set $\vz^e = (\vx^e, \vy^e)$ 

$\sim \sD_{\text{val}}$
\State Calculate $\theta_{s}^\pm$: $\theta_{s}^\pm = \theta_{s} \pm \epsilon \mathcal{L}_{\text{ce}}(\vy^e, S (\vx^e; \theta_s'))$
\State Calculate the Distillation Influence with $\vz^r, \theta_t, $

$\theta_{s}^\pm$ and  $\epsilon$: $L_{\text{influence}}$ \Comment{\cref{eqn:L_influence}}
\State Update $\theta_t$: $\theta_t \gets \theta_t - \eta_t \nabla_{\theta_t}  \mathcal{L}_t(\theta_t, \theta_s,\vz^r) $ \Comment{\cref{eqn:t_loss}}
\State Update original $\theta_s$: $\theta_s \gets \theta_s - \eta_s \nabla_{\theta_s}  \mathcal{L}_s(\theta_s, \theta_t,\vz^r) $
\State $step \gets step + 1 $

\EndWhile
\end{algorithmic}
\end{algorithm}

\begin{figure*}[t]
    \centering
    \subfigure[]{
    \begin{minipage}[t]{0.27\linewidth}
    \centering
    \includegraphics[width=\linewidth]{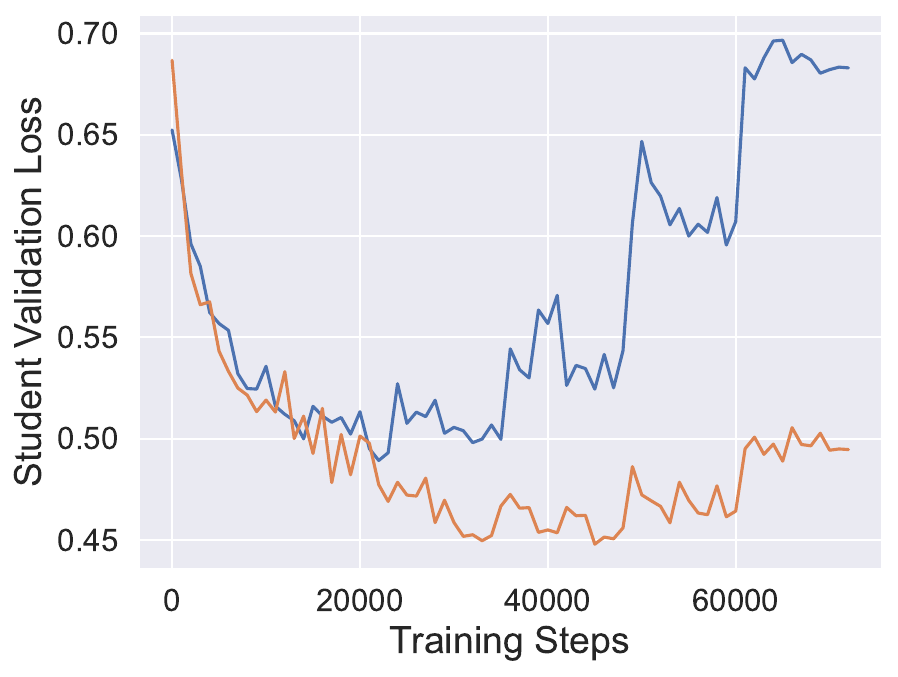}
    \end{minipage}
    }
    \subfigure[]{
    \begin{minipage}[t]{0.27\linewidth}
    \centering
    \includegraphics[width=\linewidth]{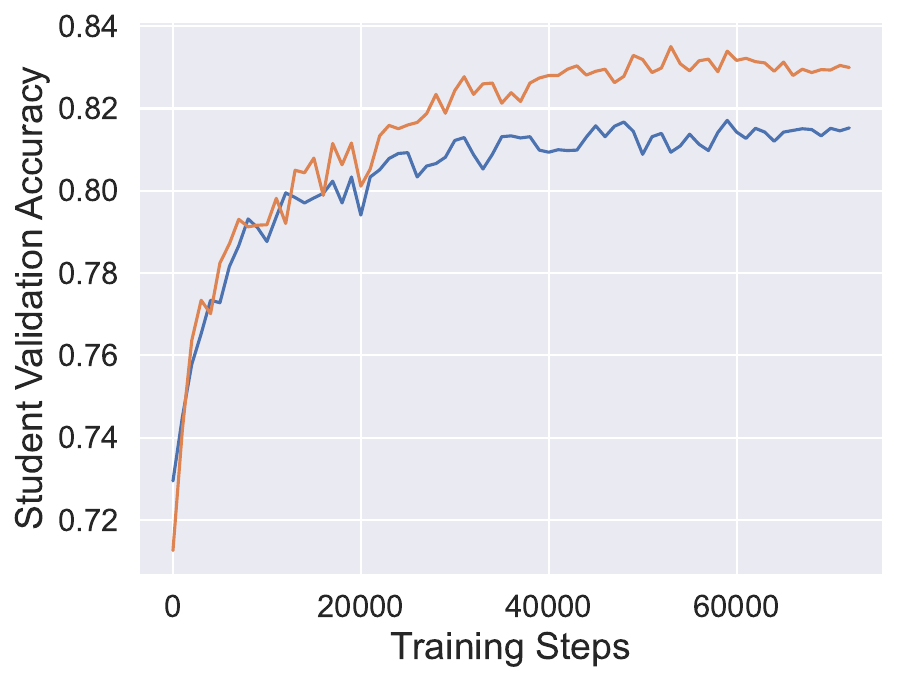}
    \end{minipage}
    }
    \subfigure[]{
    \begin{minipage}[t]{0.27\linewidth}
    \centering
    \includegraphics[width=\linewidth]{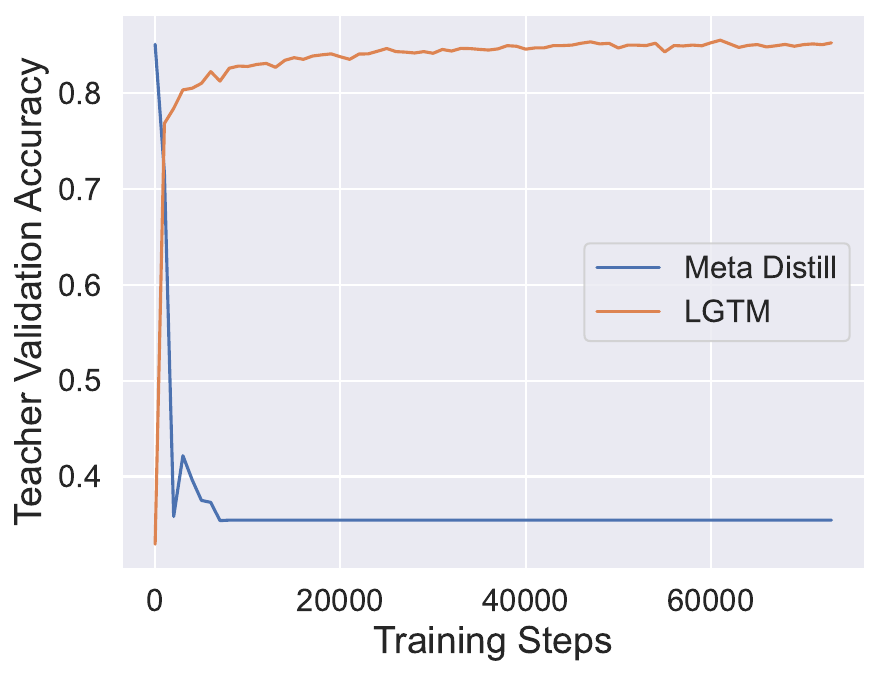}
    \end{minipage}
    }
    \vspace{-2ex}
    \caption{Performance comparison between Meta Distill~\citep{zhou2022bert} and LGTM on the MNLI validation set. We observe that for LGTM, student model does not suffer from overfitting (thanks to distillation influence), and the teacher can balance its own evolution and effective knowledge transfer (thanks to auxiliary loss).}
    \label{fig:exp_feedback}
    \vspace{-1em}
\end{figure*}

\paragraph{Finite difference approximation}
For standard neural network training, we often compute a consolidated gradient for a mini-batch of $B^r$ training samples to enhance computational efficiency.
However, in the context of determining the distillation influence for each sample, the computation of per-sample gradient $\mathcal{L}_{\text{ce}}(T(\vx_i^r;\theta_{t}^m),S(\vx_i^r;\theta_{s}^m))$ will slow down the training by a factor of $B^r$. In addition, a naive implementation is memory intensive, because it requires to keep a copy of $\nabla_{\theta_{s}}\mathcal{L}_{\text{ce}}(\vy^{e}, S (\vx^{e}; \theta_s^{m+1}))$.

To address this, we propose an efficient method for updating the teacher with the distillation influence by utilizing finite difference~\citep{gleich2005finite}, a technique commonly used in numerical analysis for approximating the derivative of a function at a given point. 
Similar to~\citep{pham2021meta,liu2018darts}, we approximate $\mathcal{L}_{\text{influence}}$ by 
\begin{equation}
\small
\begin{aligned}
\mathcal{L}_{\text{influence}} \approx \hat{\mathcal{L}}_{\text{influence}} = & \frac{1}{B^r}  \sum_{i=1}^{B^r} \bigg[\frac{ \mathcal{L}_{\text{ce}}(T(x_i;\theta_{t}^m),S(x_i;\theta_{s}^+))}{2 \epsilon}\\
& - \frac{\mathcal{L}_{\text{ce}}(T(x_i;\theta_{t}^m),S(x_i;\theta_{s}^-))}{2\epsilon}\bigg],
\label{eqn:L_influence}
\end{aligned}
\end{equation}
where $\theta_{s}^\pm = \theta_{s} \pm \epsilon \nabla \mathcal{L}_{\text{ce}}(\vy^e, S (\vx^e; \theta_s^{m+1}))$ and $\epsilon$ is a small scalar.
Our proposed method for evaluating the finite difference is computationally efficient, as it only requires two forward passes for $\theta_s$ and one backward pass for $\theta_t$ for a single batch, as opposed to a naive implementation which requires $B^r$ forward and backward passes for $\theta_{s}$ and one backward pass for $\theta_{t}$.
We provide more details of the derivation in \cref{sec:approximation}.

\paragraph{Teacher’s auxiliary loss} Inspired by~\citep{pham2021meta}, in order to balance the trade-off between self-evolution and transferability of the teacher model, we incorporate the loss with respect to the ground truth as $\mathcal{L}_{\text{aux}}$ into the final objective:
\begin{equation}
\small
\begin{aligned}
\mathcal{L}_{\text{t}}(\theta_{t}\mid \theta_{s},\vz^r) &=&& \hat{\mathcal{L}}_{\text{influence}} + \mathcal{L}_{\text{aux}},\\
\mathcal{L}_{\text{aux}} &=&& \alpha\mathcal{L}_{\text{ce}}(\vy^r,T(\vx^r;\theta_{t})) + \\ & &&(1-\alpha)\mathcal{L}_{\text{ce}}(T(\vx^r;\theta_{t}),S(\vx^r;\theta_{s}))
\label{eqn:t_loss}
\end{aligned}
\end{equation}
where $\alpha$ is the loss ratio. 

Overall, our method allows the teacher to adapt to the student's abilities and provide more personalized guidance while improving the student's generalization capability. 
We present the algorithm of LGTM in~\cref{alg:flow}.

\paragraph{Relationship with other L2T methods}
Here we interpret current learning to teach methods from the perspective of influence function. 

In the case of online distillation, it is assumed that all training samples possess an equivalent distillation influence and that the teacher model is responsible for reducing the transfer difficulty of all training samples.

In contrast, the key differentiating factor between meta distillation and online distillation is the utilization of a dynamic loss weight. We interpret this weight as a measure of the distillation influence of the current training batch $\vz^r$ on the generalization ability of the student model. Specifically, it reflects the similarity between the gradients of the training and validation batches, indicating the effect of the current training batch $\vz^r$ on the validation batch $\vz^e$ (as detailed in \cref{sec:meta_distillation}). However, it should be noted that this weight functions primarily as an adaptive learning rate, adjusting the gradient step proportionally to the degree of similarity in gradients. We illustrate the general workflow of vanilla distillation, online distillation, meta distillation and LGTM in \cref{fig:comparision}.

\begin{table*}[t]
    \small
    \centering
    \begin{tabular}{lccccccc}
    \toprule
    Model & MRPC & RTE & SST-2  & MNLI  & QNLI  & QQP & \\
    & F1/Acc. & Acc. & Acc.  & Acc. & Acc. & F1/Acc. & Avg. \\
    \midrule
Teacher \\
BERT-Base~\citep{devlin2018bert} & 89.0/85.2 & 69.5 & 93.2 & 84.3/83.9 & 91.1 & 71.5/89.2 & 84.2\\
\midrule
Student (BERT-6L)\\
KD~\citep{hinton2015distilling} & 86.7/81.4 & 64.7 & 91.2  & 81.6/80.8 & 89.0 & 70.4/88.7 & 81.6\\
PKD~\citep{sun2019patient} & 85.0/79.9 & 65.5 & 92.0   &  81.5/81.0 & 89.0 & 70.7/88.9 & 81.7\\
SKD~\citep{guo2020reducing} & 84.6/78.4 & 65.1 & 92.2   & 81.2/80.2 & 87.2 & 69.8/88.4  & 81.0\\
DIST~\citep{huang2022knowledge} & 85.8/79.8 & 65.0 & 90.9  & 81.8/80.7 & 88.0 & 70.2/88.6  & 81.2\\
TAKD~\citep{mirzadeh2020improved} & 82.4/81.7 & 64.1 & 92.5   & 82.4/81.7 & 89.4 & 70.6/88.8  & 81.6\\
RCO~\citep{jin2019knowledge} & 86.8/81.4 & 65.1 & 91.5   & 82.3/81.2 & 87.8 & 70.4/89.2  & 81.7\\
DML~\citep{zhang2018deep} & 87.5/82.8 & 64.1 & 92.4   & 82.6/81.6 & 89.5 & 70.7/88.7  & 82.2\\
ProKT~\citep{shi2020learning} & 87.1/82.3 & 65.3 &  93.0   & 82.9/82.2 & 89.5 & 71.0/89.1 & 82.5\\
PESF-KD~\citep{rao2022parameter} & 86.0/80.6 & 65.1 & 91.5  & 81.5/80.6 & 87.6 & 70.3/88.7  & 81.3\\
Meta Distill~\citep{zhou2022bert} & 85.2/79.5 & 65.6 & 92.9  & 82.4/81.4  & 88.9 & 70.1/88.5 & 81.8\\
LGTM & \textbf{88.1/83.3} & \textbf{67.4} & \textbf{93.4} & \textbf{83.4/82.5} & \textbf{90.2} & \textbf{71.7/89.3} & \textbf{83.4}\\
\bottomrule
\end{tabular}
\caption{Experimental results on the test set of GLUE (from the official test server). We bold the best results for each dataset, as well as the final average accuracy. Following ~\citep{zhou2022bert}, the student is initialized with a 6-layer pre-trained BERT~\citep{turc2019well}. We can see that LGTM outperforms all 10 baselines. }
\label{tab:main_res}
\end{table*}

\begin{figure*}[t]
    \centering
    \includegraphics[width=\linewidth]{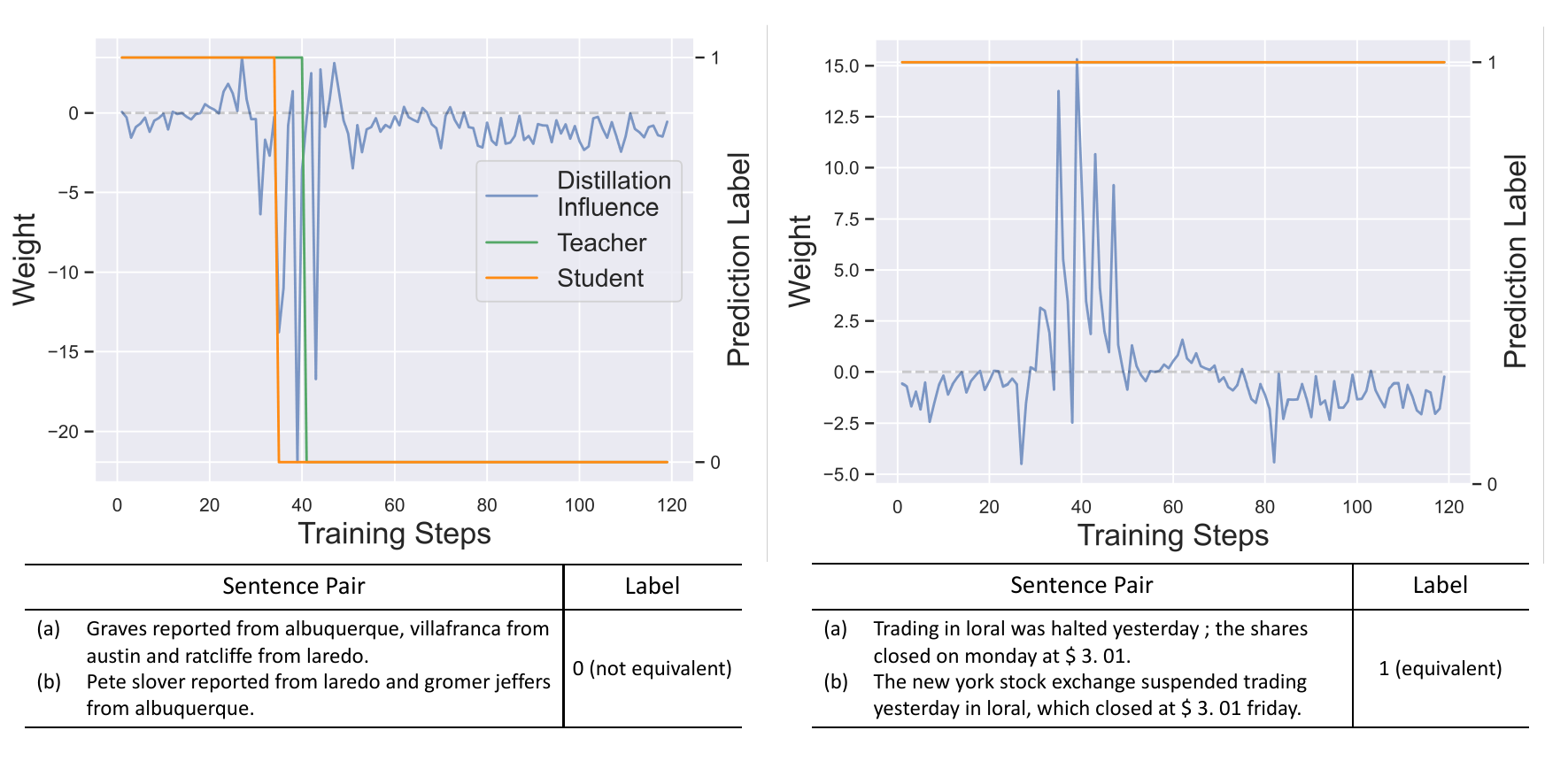}
    \caption{We select two samples in the MRPC dataset to visualize their trends of the distillation influence during training. We also visualize the relationship between the distillation influence and the predictions from the student and the teacher. Left: our method assigns negative weight to a potential difficult sample, which helps avoid overfitting. Right: our method assigns positive weight to a potential easy sample, which encourages model learning.}   
    \label{fig:exp_distill_influence1}
    \vspace{-1em}
\end{figure*}

\begin{figure}[t]
    \centering
    \subfigure{
    \begin{minipage}[t]{0.85\linewidth}
    \centering
    \includegraphics[width=\linewidth]{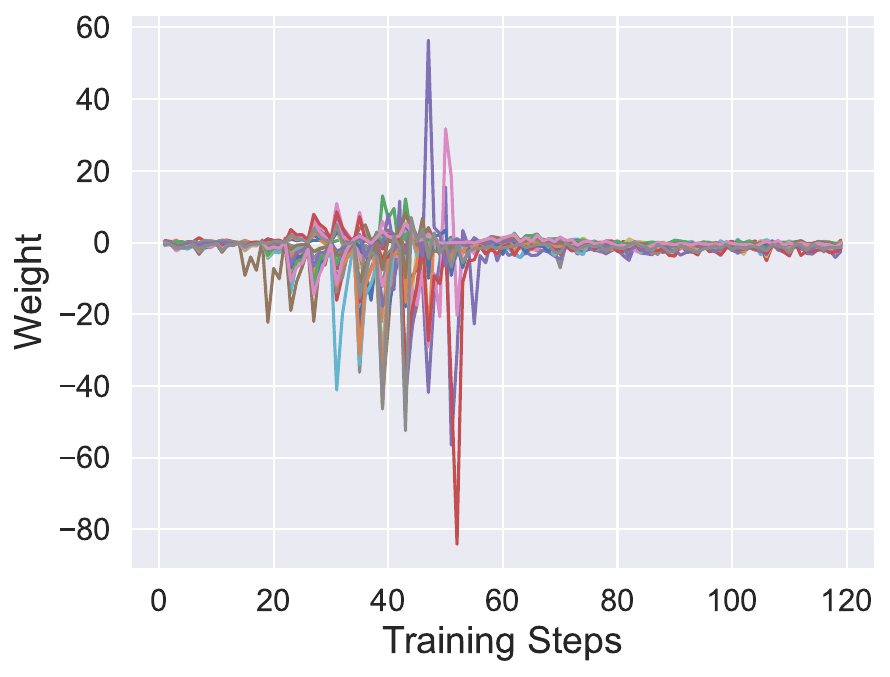}
    \end{minipage}
    }
    \subfigure{
    \begin{minipage}[t]{0.85\linewidth}
    \centering
    \includegraphics[width=\linewidth]{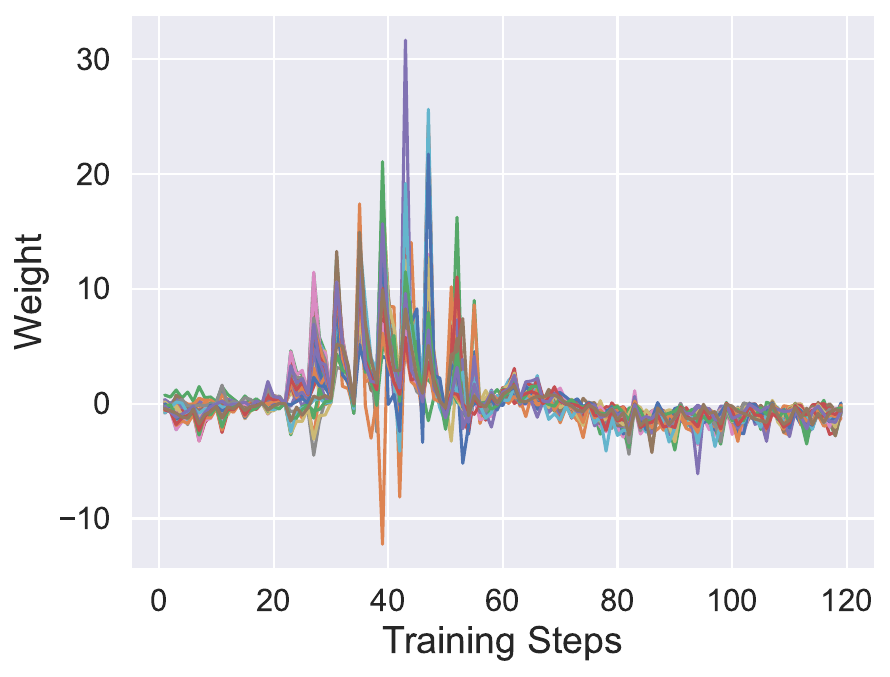}
    \end{minipage}
    }

    \caption{We visualize the trend of the distillation influence from 64 random samples in the MRPC dataset. We find that whether assigning positive or negative weight, the trend is similar. Distillation influence is usually insignificant in the beginning and end of the training, but fluctuates in the middle. We hypothese this is because our method is assigning varying weights to each sample during training, with the goal of filtering difficult samples and focusing on samples better for generalization.
}   
    \label{fig:exp_distill_influence-all}
\end{figure}

\section{Experiments}

In this section, we first describe our experiment setup including datasets and baselines in Sec.~\ref{subsec:setting}. 
Then we compare our proposed LGTM to meta distillation to gain some basic understanding of how to incorporate the student's feedback in Sec.~\ref{subsec:lgtm-vs-meta}.
To further verify the effectiveness of our method, in Sec.~\ref{subsec:sota} we compare to 10 widely adopted knowledge distillation baselines and show consistently better results.
Then we demonstrate how distillation influence works in Sec.~\ref{subsec:vis_of_di}, followed by ablation studies of LGTM in Sec.~\ref{subsec:ablation}.

\subsection{Experimental Setup}
\label{subsec:setting}
\paragraph{Datasets} We evaluate our proposed approach on text classification tasks in GLUE ~\citep{wang2018glue}: MRPC~\citep{mrpc}, RTE~\citep{wang2018glue}, SST-2~\citep{sst}, MNLI~\citep{mnli}, QNLI~\citep{rajpurkar2016squad} and QQP~\citep{chen2018quora}. For MRPC and QQP, we report both F1 and accuracy. And for other datasets, we report accuracy.

\paragraph{Baselines}
We compare our LGTM with 10 baselines: 1) KD~\citep{hinton2015distilling} 2) PKD~\citep{sun2019patient} 3) SKD~\citep{guo2020reducing} 4) DIST~\citep{huang2022knowledge}
5) TAKD~\citep{mirzadeh2020improved} 6) RCO~\citep{jin2019knowledge}
7) DML~\citep{zhang2018deep} 8) ProKT~\citep{shi2020learning} 9) PESF-KD~\citep{rao2022parameter} and 10) Meta Distill~\citep{zhou2022bert}.

\paragraph{Training setup}
Following previous works ~\citep{sun2019patient,zhou2022bert}, we distill BERT-Base~\citep{devlin2018bert} to a 6-layer BERT model. 
For all two-stage baselines, we fine-tune the models on each task. For fair comparison, both Meta Distill and LGTM utilize feedback from the validation set in the calculation of the distillation loss.
Detailed training hyperparameters can be found in \cref{sec:hyperparameters}. 

\subsection{Comparison with Meta Distillation}
\label{subsec:lgtm-vs-meta}

Given our proposed LGTM is closely related to the meta distillation line of work, here we first conduct a comparison between LGTM and a specific meta distillation method, Meta Distill~\citep{zhou2022bert}, to demonstrate the benefit of adopting distillation influence. 

We observe that for Meta Distill (blue curve) in \cref{fig:exp_feedback} (a) and (b), the validation loss of the student model gradually increases in later iterations while the validation accuracy keeps improving until a stable plateau. 
This clearly indicates that the student model is experiencing overfitting. 
One possible explanation is that excessive emphasis is placed on certain training samples that generate high loss, e.g., hard samples or outliers.
This negatively impacts the generalization ability of the student model, which leads to overfitting.

The key difference between Meta Distill and our LGTM (orange curve) is that LGTM accounts for the per-sample distillation influence while Meta Distill treats all training samples in a batch equally. 
This enables the filtering of samples that have a detrimental effect on generalization performance of the student model, leading to a steady decrease of validation loss (\cref{fig:exp_feedback} (a)) and an improved validation accuracy (\cref{fig:exp_feedback} (b)).

In terms of teacher model, it should not only impart their current knowledge to the student, but also actively seek out new information and perspectives to improve their own understanding.
As can be seen in \cref{fig:exp_feedback} (c), LGTM allows for the effective transfer of knowledge from the teacher model by incorporating the teacher auxiliary loss. The validation accuracy of the teacher model keeps improving for LGTM, but drops quickly for Meta Distill. 

\subsection{Main Results}
\label{subsec:sota}
Here we show the results of our proposed method on the test set of text classification tasks in GLUE benchmark.
As can be seen in~\cref{tab:main_res}, LGTM outperforms all 10 baselines including recent strong KD methods~\citep{guo2020reducing,huang2022knowledge,rao2022parameter,zhou2022bert}, which highlights the effectiveness of our method. 

To be more specific, our proposed method achieves state-of-the-art performance in comparison to those rely on carefully designed training pipelines or loss functions, e.g.,  PKD~\citep{sun2019patient}, SKD~\citep{guo2020reducing} and DIST~\citep{huang2022knowledge}. PKD proposes two distillation schemes, to enable the student to learn from multiple intermediate layers of the teacher model for incremental knowledge extraction. SKD and DIST both modify the form of KL-divergence loss to narrow the gap between the teacher and student models.
LGTM also does not require a series of teacher assistant models as TAKD~\citep{mirzadeh2020improved} and RCO~\citep{jin2019knowledge}.

Compared to online distillation methods, LGTM performs better than DML~\citep{zhang2018deep}, ProKT~\citep{shi2020learning} and PESF-KD~\citep{rao2022parameter}. 
This highlights the importance of incorporating student's feedback during the training process. 
An overemphasis on knowledge transfer from the training set may lead to the student overfitting the teacher's outputs, resulting in a reduction in its generalization abilities.

Furthermore, unlike meta distillation methods, e.g., Meta Distill~\citep{zhou2022bert}, our method allows for computing distillation influence of individual training samples, which enables filtering out samples that may hurt student's generalization. 
Therefore, LGTM is able to help the student to develop general understanding of the overall task while alleviate the overfitting issue.

\subsection{Analysis of Distillation Influence}
\label{subsec:vis_of_di}

We further explore the trend of the distillation influence of samples during the real training process. Here, we conduct experiments on the MRPC dataset. The task is to predict whether the sentences in a sentence pair are semantically equivalent~\citep{wang2018glue}.

First, we select two representative samples presented in \cref{fig:exp_distill_influence1} to visualize the trend of the distillation influence and its relationship between the teacher's and the student's prediction. 

On the left-side of \cref{fig:exp_distill_influence1}, we can see that during the initial stages of training, both the teacher (green) and the student (orange) have made wrong predictions. 
It might suggest that this sample poses a significant challenge for both models to learn. 
In this case, we do not want student model to mimic the output from teacher models too much because teacher model is also wrong about this sample. 
Our method is able to gradually adjust the loss weight to negative, indicating we will filter out this misleading training sample for now to make both models learn faster. 
As a result, the student model first escapes this predicament.
Then through student feedback on the validation set, the teacher model also learns to make the correct prediction.
Finally as training progresses, it is observed that both the student and the teacher are able to correctly classify this sample, resulting in the distillation influence stabilizing at a near-zero value. 

We present another example in \cref{fig:exp_distill_influence1} right, where both the student and the teacher are able to accurately predict a given sample.
It might suggest this sample is too easy for the teacher and the student. 
In this case, we want to give this sample a high positive weight to form a student-friendly decision boundary. This is similar to design a curriculum to learn from easy samples to hard ones in curriculum learning~\citep{Soviany2022curriculum}.

We also visualize an average trend of distillation influence in \cref{fig:exp_distill_influence-all}, based on 64 samples that are randomly chosen from MRPC. 
We observe that the distillation influence is usually insignificant in the beginning and end of the training, but fluctuates in the middle. 
This is reasonable since our method is assigning varying weights to each sample during training, with the goal of filtering difficult samples and focusing on samples better for generalization. 

\begin{table}[t]
    \centering
    \resizebox{\linewidth}{!}{
    \begin{tabular}{lccccccc}
    \toprule
    Model & MRPC & RTE & SST-2  & MNLI  & QNLI  & QQP & \\
    & F1/Acc. & Acc. & Acc.  & Acc. & Acc. & F1/Acc.& Avg.  \\
    \midrule
    Teacher \\
BERT-Base~\citep{devlin2018bert} & 89.0/85.2 & 69.5 & 93.2  & 84.3/83.9 & 91.1 & 71.5/89.2 & 85.4 \\
\midrule
    Student (BERT-6L)\\
    DIST~\citep{huang2022knowledge} & 85.8/79.8 & 65.0 & 90.9  & 81.8/80.7 & 88.0 & 70.2/88.6 & 81.2 \\
    LGTM (w. DIST) & \textbf{88.3/83.5} & \textbf{67.7} & \textbf{91.7} & \textbf{82.5}/\textbf{80.8}   & \textbf{90.4} & \textbf{71.0}/\textbf{88.9} & \textbf{82.9} \\
    \midrule
    Student (BERT-6L)\\
    MSE & 85.7/80.1 & 65.1 & 91.3 & 82.0/81.6 & 88.7 & 71.3/89.0 & 81.7 \\
    LGTM (w. MSE) & \textbf{88.1}/\textbf{83.7} & \textbf{65.8} & \textbf{92.4} & \textbf{82.5}/\textbf{80.8}  & \textbf{89.9} & \textbf{71.6}/\textbf{89.2} & \textbf{82.7} \\
    \bottomrule
    \end{tabular}}
    \caption{Experimental results on the test set of GLUE when training with different KD objectives. Our LGTM consistently beats the original methods, which validates the compatibility of LGTM to these losses.}
    \label{tab:loss-abl}
\end{table}

\subsection{Ablation Study}
\label{subsec:ablation}
Given limited space, we present three studies in this section and show more ablation studies in~\cref{sec:more_ablations}.

\paragraph{Finite difference approximation}
Recall in~\cref{sec:methods}, we introduce finite difference approximation (FDA) for estimating the distillation influence of each sample. It is designed to address the slowness of computing per-sample gradients. 
As shown in \cref{tab:comparsion_fda}, here we conduct an ablation experiment on the MRPC dataset to evaluate its usefulness.
We show that with FDA, our method only requires 11 minutes to complete the training, while the naive training without FDA requires 117 minutes.
Such a significant reduction in training time (i.e., more than $10 \times$ speedup) highlights the computational efficiency of the proposed FDA technique.
Furthermore, we assess the performance on the validation set of the MRPC dataset and observe that training with FDA result in an F1 score of 90.4, while training without FDA resulted in a score of 90.7. There is only a slight drop in performance with the approximation.
\begin{table}[]
    \small
    \centering
    \begin{tabular}{ccc}
    \toprule
         & Training time & F1 \\
    \midrule
     LGTM w/o FDA    & 117min & 90.7 \\
     LGTM w/ FDA    & 11min & 90.4 \\
     \bottomrule
    \end{tabular}
    \caption{The comparsion of LGTM with FDA and without FDA method. While their performance are similar, LGTM with FDA is 10$\times$ faster than without it.}
    \label{tab:comparsion_fda}
\end{table}

\paragraph{Distillation loss}
There are other distillation losses in the context of knowledge distillation. 
Here we want to evaluate whether LGTM can adapt to those objectives.
In particular, we consider the modified loss used in DIST~\citep{huang2022knowledge} and the common mean squared error (MSE).
As can be seen in \cref{tab:loss-abl}, our LGTM consistently beats the original methods that utilize these distillation objectives, which validates the compatibility of LGTM to different distillation objectives.

\paragraph{Student model size}
Here we conduct experiments to evaluate the performance of our proposed method in scenarios where there is a larger capacity difference between the teacher and student models. 
Specifically, we perform knowledge distillation from a BERT-Base model~\citep{devlin2018bert} to a 4-layer BERT model.
As can be seen from~\cref{tab:stu_size-abl}, LGTM consistently outperforms other baselines in most of the tasks except competitive results on SST-2. 
This indicates the robustness of our method which suggests its wide usage in various knowledge distillation settings.

\section{Related Work}
The core of knowledge distillation ~\citep{hinton2015distilling} relies on how to formulate and transfer the knowledge from the teacher to student. Three key aspects are typically considered: the teacher model from which knowledge is transferred (learning target), the data on which the model is trained (learning material), and the objective function that defines the learning objective. Efforts have been made to make knowledge distillation more student-friendly by reducing the difficulties in these aspects\cite{li2021dynamic}. 

On learning target, \citet{jin2019knowledge,mirzadeh2020improved} introduce teacher assistant models of intermediate timestep or training time step respectively to narrow the gap between the teacher and student models.  \citet{park2021learning,shi2020learning} propose updating the teacher and student jointly to make the teacher aware of the student's state. \citet{rao2022parameter} trains for more timestep to smooth the distribution of the teacher for a easier transfer. 

In terms of learning material, TinyBERT~\citep{jiao2020tinybert} suggests augmenting the training data to make it more diverse. \citet{kim2022tutoring} proposes training the student with samples that are easy for the teacher but difficult for the student. 
With respect to learning objective, the most common approach is to match the probabilistic prediction scores of the teacher and student models using KL-divergence. However, this can cause problems during training, leading to poor performance. \citet{guo2020reducing,huang2022knowledge} propose to soft the constraint by a more tolerated loss. \citet{pham2021meta,zhou2022bert} propose using the student's performance as the optimization objective for the teacher model, allowing the teacher to optimize its knowledge transfer based on feedback from the student. \citet{wang2022improved} proposes to select the appropriate knowledge to guide the optimization of the student.

\begin{table}[t]
    \centering
    \resizebox{\linewidth}{!}{
    \begin{tabular}{lccccccc}
    \toprule
    Model & MRPC & RTE & SST-2  & MNLI  & QNLI  & QQP & \\
    & F1/Acc. & Acc. & Acc.  & Acc. & Acc. & F1/Acc. & Avg. \\
    \midrule
    Teacher \\
BERT-Base~\citep{devlin2018bert} & 89.0/85.2 & 69.5 & 93.2  & 84.3/83.9 & 91.1 & 71.5/89.2 & 84.2 \\
\midrule
Student (BERT-4L)\\
KD~\citep{hinton2015distilling} & 85.6/79.9 & 63.6 & 90.4 & 80.5/79.9 & 87.7 & 70.2/88.5 & 80.2\\
PKD~\citep{sun2019patient} & 84.6/79.5 & 63.5 & 90.0 & 80.1/79.0 & 86.9 & 69.1/88.5 & 79.7 \\
DIST~\citep{huang2022knowledge} & 84.3/78.6 & 63.6 & 90.1 & 80.3/79.4 & 87.6 & 70.1/88.5 & 79.8 \\
SKD~\citep{guo2020reducing} & 83.4/77.7 & 62.0 & 89.1 & 77.9/77.2 & 86.9 & 68.2/86.9 & 78.5 \\
ProKT~\citep{shi2020learning} & 86.7/80.9 & 64.0 & 90.3 & 81.6/80.4 & 88.0 & 70.7/88.8 & 80.7 \\
Meta Distill~\citep{zhou2022bert} & 84.6/78.8 & 64.8 & \textbf{90.9} & 80.5/79.2 & 87.5 & 69.7/88.4 & 80.2 \\
LGTM & \textbf{87.4/82.3} & \textbf{65.2} & 90.8 & \textbf{81.8/80.3} & \textbf{88.9} & \textbf{71.1/88.9} & \textbf{81.4} \\
    \bottomrule
    \end{tabular}}
    \caption{Experimental results on the test set of GLUE. The student is initialized with a 4-layer pre-trained BERT~\citep{turc2019well}. LGTM again outperforms strong baselines when the capacity gap is larger between teacher and student.}
    \label{tab:stu_size-abl}
    \vspace{-1em}
\end{table}

\section{Conclusion}

In this paper, we first revisit several learning to teach paradigms in knowledge distillation. 
Then we propose distillation influence to determine how distilling from each training sample impacts the student’s generalization ability. 
By visualizing how the distillation influence of each sample changes during training, we can see that a simple re-weighting using distillation influence is able to help student training, e.g., reduce overfitting.
Built on top of distillation influence, we propose our learning to teach framework, LGTM, that consistently outperforms existing knowledge distillation methods on text classification tasks in the GLUE benchmark.

\section*{Limitations}
Although LGTM has demonstrated superior performance in task-specific knowledge distillation, it is worth investigating the potential benefits of combining LGTM with pre-training knowledge distillation~\citep{jiao2020tinybert,wang2020minilm}. Additionally, while our experiments have been limited to text classification tasks, which are relatively simple for current pre-trained language models, future work should explore the application of LGTM to more complex text generation tasks. 

\section*{Ethics Statement}
During the training process, the teacher and student models are initialized from pre-trained models. However, pre-trained language models are vulnerable to potential ethical and social risk as mentioned by~\citet{bommasani2021opportunities} and \citet{weidinger2021ethical}.
Therefore, the teacher and student models can be exposed  to similar social risks of large language models.

\section*{Acknowledgements}
We thank Yongfei Liu and Zhengkun Zhang for their insightful discussion and the anonymous reviewers for their helpful comments. This work was supported by the National Key R\&D Program of China (2022YFB4701400/4701402), SZSTC Grant (JCYJ20190809172201639, WDZC20200820200-\\655001), Shenzhen Key Laboratory (ZDSYS2021-\\0623092001004),  and Beijing Key Lab of Networked Multimedia.

\bibliography{anthology,custom}

\begin{thebibliography}{49}
\expandafter\ifx\csname natexlab\endcsname\relax\def\natexlab#1{#1}\fi

\bibitem[{Bommasani et~al.(2021)Bommasani, Hudson, Adeli, Altman, Arora, von Arx, Bernstein, Bohg, Bosselut, Brunskill et~al.}]{bommasani2021opportunities}
Rishi Bommasani, Drew~A Hudson, Ehsan Adeli, Russ Altman, Simran Arora, Sydney von Arx, Michael~S Bernstein, Jeannette Bohg, Antoine Bosselut, Emma Brunskill, et~al. 2021.
\newblock On the opportunities and risks of foundation models.
\newblock \emph{arXiv preprint arXiv:2108.07258}.

\bibitem[{Chen et~al.(2018)Chen, Zhang, Zhang, and Zhao}]{chen2018quora}
Zihan Chen, Hongbo Zhang, Xiaoji Zhang, and Leqi Zhao. 2018.
\newblock Quora question pairs.

\bibitem[{Cho and Hariharan(2019)}]{cho2019efficacy}
Jang~Hyun Cho and Bharath Hariharan. 2019.
\newblock On the efficacy of knowledge distillation.
\newblock In \emph{Proceedings of the IEEE/CVF international conference on computer vision}, pages 4794--4802.

\bibitem[{Dai et~al.(2019)Dai, Yang, Yang, Carbonell, Le, and Salakhutdinov}]{dai2019transformer}
Zihang Dai, Zhilin Yang, Yiming Yang, Jaime~G Carbonell, Quoc Le, and Ruslan Salakhutdinov. 2019.
\newblock Transformer-xl: Attentive language models beyond a fixed-length context.
\newblock In \emph{Proceedings of the 57th Annual Meeting of the Association for Computational Linguistics}, pages 2978--2988.

\bibitem[{Devlin et~al.(2019)Devlin, Chang, Lee, and Toutanova}]{devlin2018bert}
Jacob Devlin, Ming{-}Wei Chang, Kenton Lee, and Kristina Toutanova. 2019.
\newblock {BERT:} pre-training of deep bidirectional transformers for language understanding.
\newblock In \emph{{NAACL-HLT} {(1)}}, pages 4171--4186. Association for Computational Linguistics.

\bibitem[{Dolan and Brockett(2005)}]{mrpc}
William~B. Dolan and Chris Brockett. 2005.
\newblock Automatically constructing a corpus of sentential paraphrases.
\newblock In \emph{IWP@IJCNLP}.

\bibitem[{Fan et~al.(2019)Fan, Grave, and Joulin}]{fan2019reducing}
Angela Fan, Edouard Grave, and Armand Joulin. 2019.
\newblock Reducing transformer depth on demand with structured dropout.
\newblock In \emph{International Conference on Learning Representations}.

\bibitem[{Fan et~al.(2018)Fan, Tian, Qin, Li, and Liu}]{fan2018learning}
Yang Fan, Fei Tian, Tao Qin, Xiang-Yang Li, and Tie-Yan Liu. 2018.
\newblock Learning to teach.
\newblock In \emph{International Conference on Learning Representations}.

\bibitem[{Ghorbani and Zou(2019)}]{ghorbani2019data}
Amirata Ghorbani and James Zou. 2019.
\newblock Data shapley: Equitable valuation of data for machine learning.
\newblock In \emph{International Conference on Machine Learning}, pages 2242--2251. PMLR.

\bibitem[{Gleich(2005)}]{gleich2005finite}
David Gleich. 2005.
\newblock Finite calculus: A tutorial for solving nasty sums.
\newblock \emph{Stanford University}.

\bibitem[{Guo et~al.(2022)Guo, Chen, Hu, Zhu, He, and Cai}]{guo2020reducing}
Jia Guo, Minghao Chen, Yao Hu, Chen Zhu, Xiaofei He, and Deng Cai. 2022.
\newblock Reducing the teacher-student gap via spherical knowledge disitllation.
\newblock \emph{openreview.net}.

\bibitem[{Hara et~al.(2019)Hara, Nitanda, and Maehara}]{hara2019data}
Satoshi Hara, Atsushi Nitanda, and Takanori Maehara. 2019.
\newblock Data cleansing for models trained with sgd.
\newblock \emph{Advances in Neural Information Processing Systems}, 32.

\bibitem[{Hinton et~al.(2015)Hinton, Vinyals, Dean et~al.}]{hinton2015distilling}
Geoffrey Hinton, Oriol Vinyals, Jeff Dean, et~al. 2015.
\newblock Distilling the knowledge in a neural network.
\newblock \emph{arXiv preprint arXiv:1503.02531}, 2(7).

\bibitem[{Huang et~al.(2022)Huang, You, Wang, Qian, and Xu}]{huang2022knowledge}
Tao Huang, Shan You, Fei Wang, Chen Qian, and Chang Xu. 2022.
\newblock Knowledge distillation from a stronger teacher.
\newblock \emph{Advances in Neural Information Processing Systems}.

\bibitem[{Jia et~al.(2019)Jia, Dao, Wang, Hubis, Hynes, G{\"u}rel, Li, Zhang, Song, and Spanos}]{jia2019towards}
Ruoxi Jia, David Dao, Boxin Wang, Frances~Ann Hubis, Nick Hynes, Nezihe~Merve G{\"u}rel, Bo~Li, Ce~Zhang, Dawn Song, and Costas~J Spanos. 2019.
\newblock Towards efficient data valuation based on the shapley value.
\newblock In \emph{The 22nd International Conference on Artificial Intelligence and Statistics}, pages 1167--1176. PMLR.

\bibitem[{Jiao et~al.(2020)Jiao, Yin, Shang, Jiang, Chen, Li, Wang, and Liu}]{jiao2020tinybert}
Xiaoqi Jiao, Yichun Yin, Lifeng Shang, Xin Jiang, Xiao Chen, Linlin Li, Fang Wang, and Qun Liu. 2020.
\newblock Tinybert: Distilling bert for natural language understanding.
\newblock In \emph{Findings of the Association for Computational Linguistics: EMNLP 2020}, pages 4163--4174.

\bibitem[{Jin et~al.(2019)Jin, Peng, Wu, Liu, Liu, Liang, Yan, and Hu}]{jin2019knowledge}
Xiao Jin, Baoyun Peng, Yichao Wu, Yu~Liu, Jiaheng Liu, Ding Liang, Junjie Yan, and Xiaolin Hu. 2019.
\newblock Knowledge distillation via route constrained optimization.
\newblock In \emph{Proceedings of the IEEE/CVF International Conference on Computer Vision}, pages 1345--1354.

\bibitem[{Kim et~al.(2022)Kim, Park, Lee, Mok, Choi, and Lee}]{kim2022tutoring}
Junho Kim, Jun-Hyung Park, Mingyu Lee, Wing-Lam Mok, Joon-Young Choi, and SangKeun Lee. 2022.
\newblock Tutoring helps students learn better: Improving knowledge distillation for bert with tutor network.
\newblock In \emph{Proceedings of the 2022 Conference on Empirical Methods in Natural Language Processing}, pages 7371--7382.

\bibitem[{Kim et~al.(2021)Kim, Gholami, Yao, Mahoney, and Keutzer}]{kim2021bert}
Sehoon Kim, Amir Gholami, Zhewei Yao, Michael~W Mahoney, and Kurt Keutzer. 2021.
\newblock I-bert: Integer-only bert quantization.
\newblock In \emph{International conference on machine learning}, pages 5506--5518. PMLR.

\bibitem[{Koh and Liang(2017)}]{koh2017understanding}
Pang~Wei Koh and Percy Liang. 2017.
\newblock Understanding black-box predictions via influence functions.
\newblock In \emph{International conference on machine learning}, pages 1885--1894. PMLR.

\bibitem[{Li et~al.(2021{\natexlab{a}})Li, Cotterell, and Sachan}]{li2021differentiable}
Jiaoda Li, Ryan Cotterell, and Mrinmaya Sachan. 2021{\natexlab{a}}.
\newblock Differentiable subset pruning of transformer heads.
\newblock \emph{Transactions of the Association for Computational Linguistics}, 9:1442--1459.

\bibitem[{Li et~al.(2021{\natexlab{b}})Li, Lin, Ren, Li, Zhou, and Sun}]{li2021dynamic}
Lei Li, Yankai Lin, Shuhuai Ren, Peng Li, Jie Zhou, and Xu~Sun. 2021{\natexlab{b}}.
\newblock Dynamic knowledge distillation for pre-trained language models.
\newblock In \emph{Proceedings of the 2021 Conference on Empirical Methods in Natural Language Processing}, pages 379--389.

\bibitem[{Liu et~al.(2018)Liu, Simonyan, and Yang}]{liu2018darts}
Hanxiao Liu, Karen Simonyan, and Yiming Yang. 2018.
\newblock Darts: Differentiable architecture search.
\newblock In \emph{International Conference on Learning Representations}.

\bibitem[{Liu et~al.(2019)Liu, Ott, Goyal, Du, Joshi, Chen, Levy, Lewis, Zettlemoyer, and Stoyanov}]{liu2019roberta}
Yinhan Liu, Myle Ott, Naman Goyal, Jingfei Du, Mandar Joshi, Danqi Chen, Omer Levy, Mike Lewis, Luke Zettlemoyer, and Veselin Stoyanov. 2019.
\newblock Roberta: A robustly optimized bert pretraining approach.
\newblock \emph{arXiv preprint arXiv:1907.11692}.

\bibitem[{Mirzadeh et~al.(2020)Mirzadeh, Farajtabar, Li, Levine, Matsukawa, and Ghasemzadeh}]{mirzadeh2020improved}
Seyed~Iman Mirzadeh, Mehrdad Farajtabar, Ang Li, Nir Levine, Akihiro Matsukawa, and Hassan Ghasemzadeh. 2020.
\newblock Improved knowledge distillation via teacher assistant.
\newblock In \emph{Proceedings of the AAAI conference on artificial intelligence}, volume~34, pages 5191--5198.

\bibitem[{Park et~al.(2021)Park, Cha, Kim, Han et~al.}]{park2021learning}
Dae~Young Park, Moon-Hyun Cha, Daesin Kim, Bohyung Han, et~al. 2021.
\newblock Learning student-friendly teacher networks for knowledge distillation.
\newblock \emph{Advances in Neural Information Processing Systems}, 34:13292--13303.

\bibitem[{Pham et~al.(2021)Pham, Dai, Xie, and Le}]{pham2021meta}
Hieu Pham, Zihang Dai, Qizhe Xie, and Quoc~V Le. 2021.
\newblock Meta pseudo labels.
\newblock In \emph{Proceedings of the IEEE/CVF Conference on Computer Vision and Pattern Recognition}, pages 11557--11568.

\bibitem[{Pruthi et~al.(2020)Pruthi, Liu, Kale, and Sundararajan}]{pruthi2020estimating}
Garima Pruthi, Frederick Liu, Satyen Kale, and Mukund Sundararajan. 2020.
\newblock Estimating training data influence by tracing gradient descent.
\newblock \emph{Advances in Neural Information Processing Systems}, 33:19920--19930.

\bibitem[{Rajpurkar et~al.(2016)Rajpurkar, Zhang, Lopyrev, and Liang}]{rajpurkar2016squad}
Pranav Rajpurkar, Jian Zhang, Konstantin Lopyrev, and Percy Liang. 2016.
\newblock Squad: 100, 000+ questions for machine comprehension of text.
\newblock In \emph{EMNLP}.

\bibitem[{Rao et~al.(2022)Rao, Meng, Ding, Qi, and Tao}]{rao2022parameter}
Jun Rao, Xv~Meng, Liang Ding, Shuhan Qi, and Dacheng Tao. 2022.
\newblock Parameter-efficient and student-friendly knowledge distillation.
\newblock \emph{arXiv preprint arXiv:2205.15308}.

\bibitem[{Shi et~al.(2020)Shi, Song, Zhou, Li, and Li}]{shi2020learning}
Wenxian Shi, Yuxuan Song, Hao Zhou, Bohan Li, and Lei Li. 2020.
\newblock Learning from deep model via exploring local targets.
\newblock \emph{openreview.net}.

\bibitem[{Socher et~al.(2013)Socher, Perelygin, Wu, Chuang, Manning, Ng, and Potts}]{sst}
Richard Socher, Alex Perelygin, Jean Wu, Jason Chuang, Christopher~D. Manning, Andrew~Y. Ng, and Christopher Potts. 2013.
\newblock Recursive deep models for semantic compositionality over a sentiment treebank.
\newblock In \emph{{EMNLP}}.

\bibitem[{Soviany et~al.(2022)Soviany, Ionescu, Rota, and Sebe}]{Soviany2022curriculum}
Petru Soviany, Radu~Tudor Ionescu, Paolo Rota, and Nicu Sebe. 2022.
\newblock Curriculum learning: A survey.
\newblock \emph{International Journal of Computer Vision}.

\bibitem[{Stanton et~al.(2021)Stanton, Izmailov, Kirichenko, Alemi, and Wilson}]{stanton2021does}
Samuel Stanton, Pavel Izmailov, Polina Kirichenko, Alexander~A Alemi, and Andrew~G Wilson. 2021.
\newblock Does knowledge distillation really work?
\newblock \emph{Advances in Neural Information Processing Systems}, 34:6906--6919.

\bibitem[{Sun et~al.(2019)Sun, Cheng, Gan, and Liu}]{sun2019patient}
Siqi Sun, Yu~Cheng, Zhe Gan, and Jingjing Liu. 2019.
\newblock Patient knowledge distillation for bert model compression.
\newblock In \emph{Proceedings of the 2019 Conference on Empirical Methods in Natural Language Processing and the 9th International Joint Conference on Natural Language Processing (EMNLP-IJCNLP)}, pages 4323--4332.

\bibitem[{Sun et~al.(2020)Sun, Yu, Song, Liu, Yang, and Zhou}]{sun2020mobilebert}
Zhiqing Sun, Hongkun Yu, Xiaodan Song, Renjie Liu, Yiming Yang, and Denny Zhou. 2020.
\newblock Mobilebert: a compact task-agnostic bert for resource-limited devices.
\newblock In \emph{Proceedings of the 58th Annual Meeting of the Association for Computational Linguistics}, pages 2158--2170.

\bibitem[{Tang et~al.(2019)Tang, Lu, Liu, Mou, Vechtomova, and Lin}]{tang2019distilling}
Raphael Tang, Yao Lu, Linqing Liu, Lili Mou, Olga Vechtomova, and Jimmy Lin. 2019.
\newblock Distilling task-specific knowledge from bert into simple neural networks.
\newblock \emph{arXiv preprint arXiv:1903.12136}.

\bibitem[{Turc et~al.(2019)Turc, Chang, Lee, and Toutanova}]{turc2019well}
Iulia Turc, Ming-Wei Chang, Kenton Lee, and Kristina Toutanova. 2019.
\newblock Well-read students learn better: On the importance of pre-training compact models.

\bibitem[{Wang et~al.(2018)Wang, Singh, Michael, Hill, Levy, and Bowman}]{wang2018glue}
Alex Wang, Amanpreet Singh, Julian Michael, Felix Hill, Omer Levy, and Samuel~R Bowman. 2018.
\newblock Glue: A multi-task benchmark and analysis platform for natural language understanding.
\newblock In \emph{International Conference on Learning Representations}.

\bibitem[{Wang et~al.(2022{\natexlab{a}})Wang, Yang, Huang, Song, and Huang}]{wang2022efficient}
Chaofei Wang, Qisen Yang, Rui Huang, Shiji Song, and Gao Huang. 2022{\natexlab{a}}.
\newblock Efficient knowledge distillation from model checkpoints.
\newblock In \emph{Advances in Neural Information Processing Systems}.

\bibitem[{Wang et~al.(2022{\natexlab{b}})Wang, Lu, Mu, Hu, Xiao, and Zhu}]{wang2022improved}
Chenglong Wang, Yi~Lu, Yongyu Mu, Yimin Hu, Tong Xiao, and Jingbo Zhu. 2022{\natexlab{b}}.
\newblock Improved knowledge distillation for pre-trained language models via knowledge selection.
\newblock In \emph{Findings of the Association for Computational Linguistics: EMNLP 2022}, pages 6232--6244.

\bibitem[{Wang et~al.(2020)Wang, Wei, Dong, Bao, Yang, and Zhou}]{wang2020minilm}
Wenhui Wang, Furu Wei, Li~Dong, Hangbo Bao, Nan Yang, and Ming Zhou. 2020.
\newblock Minilm: Deep self-attention distillation for task-agnostic compression of pre-trained transformers.
\newblock \emph{Advances in Neural Information Processing Systems}, 33:5776--5788.

\bibitem[{Weidinger et~al.(2021)Weidinger, Mellor, Rauh, Griffin, Uesato, Huang, Cheng, Glaese, Balle, Kasirzadeh et~al.}]{weidinger2021ethical}
Laura Weidinger, John Mellor, Maribeth Rauh, Conor Griffin, Jonathan Uesato, Po-Sen Huang, Myra Cheng, Mia Glaese, Borja Balle, Atoosa Kasirzadeh, et~al. 2021.
\newblock Ethical and social risks of harm from language models.
\newblock \emph{arXiv preprint arXiv:2112.04359}.

\bibitem[{Williams et~al.(2018)Williams, Nangia, and Bowman}]{mnli}
Adina Williams, Nikita Nangia, and Samuel~R. Bowman. 2018.
\newblock A broad-coverage challenge corpus for sentence understanding through inference.
\newblock In \emph{{NAACL-HLT}}.

\bibitem[{Yang et~al.(2019)Yang, Dai, Yang, Carbonell, Salakhutdinov, and Le}]{yang2019xlnet}
Zhilin Yang, Zihang Dai, Yiming Yang, Jaime Carbonell, Russ~R Salakhutdinov, and Quoc~V Le. 2019.
\newblock Xlnet: Generalized autoregressive pretraining for language understanding.
\newblock \emph{Advances in neural information processing systems}, 32.

\bibitem[{Zhang et~al.(2020)Zhang, Hou, Yin, Shang, Chen, Jiang, and Liu}]{zhang2020ternarybert}
Wei Zhang, Lu~Hou, Yichun Yin, Lifeng Shang, Xiao Chen, Xin Jiang, and Qun Liu. 2020.
\newblock Ternarybert: Distillation-aware ultra-low bit bert.
\newblock In \emph{Proceedings of the 2020 Conference on Empirical Methods in Natural Language Processing (EMNLP)}, pages 509--521.

\bibitem[{Zhang et~al.(2018)Zhang, Xiang, Hospedales, and Lu}]{zhang2018deep}
Ying Zhang, Tao Xiang, Timothy~M Hospedales, and Huchuan Lu. 2018.
\newblock Deep mutual learning.
\newblock In \emph{Proceedings of the IEEE conference on computer vision and pattern recognition}, pages 4320--4328.

\bibitem[{Zhou et~al.(2022)Zhou, Xu, and McAuley}]{zhou2022bert}
Wangchunshu Zhou, Canwen Xu, and Julian McAuley. 2022.
\newblock Bert learns to teach: Knowledge distillation with meta learning.
\newblock In \emph{Proceedings of the 60th Annual Meeting of the Association for Computational Linguistics (Volume 1: Long Papers)}, pages 7037--7049.

\bibitem[{Zhu et~al.(2018)Zhu, Gong et~al.}]{zhu2018knowledge}
Xiatian Zhu, Shaogang Gong, et~al. 2018.
\newblock Knowledge distillation by on-the-fly native ensemble.
\newblock \emph{Advances in neural information processing systems}, 31.

\end{thebibliography}
\bibliographystyle{acl_natbib}

\appendix

\section{The Derivation of Distillation Influence}
\label{sec:influence}
As described by~\citet{pruthi2020estimating}, the influence of a training sample $z=(x,y)$ on a test sample $z'=(x',y')$ can be traced by examining the change in loss of model $w$ on the test sample. The influence function is defined as the total reduction in loss on the test sample $z'$ induced by the training process whenever the training sample $z$ is utilized:
\begin{equation}
\small
    \mathcal{I}(z,z') = \sum_{t:z_t=z} \mathcal{L}(w_t,z')-\mathcal{L}(w_{t+1},z').
\label{equ:tracin}
\end{equation}
where $w_{t+1}=w_t -\eta_{w}\mathcal{L}(w_t,z)$ and $\eta_{w}$ is the learning rate and the model are parameterized by $w_t$ and $w_{t+1}$.

In this context, we will focus on the influence of the current training batch on the student model's performance on the validation data. To improve computation efficiency, a batch of samples is drawn from the validation set to evaluate the model's generalization performance. As a result, the influence on a single validation sample, as described in \cref{equ:tracin}, is extended to a batch of validation samples $\vz^e$. The influence of the current training batch $\vz^{r}$ on the validation batch $\vz^{e}$ is defined as follows:
\begin{equation}
\small
\begin{aligned}
 &\mathcal{I}(\vz^r,\vz^e)= \mathcal{L}_{\text{val}}(\theta_{s}^{m},\vz^{e})-\mathcal{L}_{\text{val}}(\theta_{s}^{m+1},\vz^{e}) \\
    &= \mathcal{L}_{\text{ce}}(\vy^{e}, S (\vx^{e}; \theta_s^{m}))-\mathcal{L}_{\text{ce}}(\vy^{e}, S (\vx^{e}; \theta_s^{m+1})),
\end{aligned}
\end{equation}
where $\theta_{s}^{m+1}=\theta_{s}^{m}-\eta_{s}\mathcal{L}_{\text{s}}(\theta_{s}^m, \theta_{t}^m,\vz^r)$.

By applying the Taylor expansion, we can approximate $\mathcal{L}_{\text{val}}(\theta_{s}^{m},\vz^{e})$ as follows:
\begin{equation}
\small
\begin{aligned}
&\mathcal{L}_{\text{val}}(\theta_{s}^{m},\vz^{e})=\mathcal{L}_{\text{val}}(\theta_{s}^{m+1},\vz^{e})+(\theta_{s}^{m}-\theta_{s}^{m+1})^\intercal \\
&\nabla_{\theta_{s}} \mathcal{L}_{\text{val}}(\theta_{s}^{m+1}, \vz^e)+O(||\theta_{s}^{m}-\theta_{s}^{m+1}||^2) \\
&\approx \mathcal{L}_{\text{val}}(\theta_{s}^{m+1},\vz^{e})+(\eta_{s}\nabla_{\theta_{s}}\mathcal{L}_{s}(\theta_{s}^m, \theta_{t}^m, \vz^r))^\intercal\\&\nabla_{\theta_{s}}  \mathcal{L}_{\text{val}}(\theta_{s}^{m+1}, \vz^e) \\
\end{aligned}
\end{equation}
As a result, we approximate the $\mathcal{I}(\vz^{r},\vz^{e})$ as follows:
\begin{equation}
\small
\begin{aligned}
&\mathcal{L}_{\text{val}}(\theta_{s}^{m},\vz^{e})-\mathcal{L}_{\text{val}}(\theta_{s}^{m+1},\vz^{e}) \\
&\approx (\eta_{s}\nabla_{\theta_{s}}\mathcal{L}_{s}(\theta_{s}^m, \theta_{t}^m, \vz^r))^\intercal\nabla_{\theta_{s}} \mathcal{L}_{\text{val}}(\theta_{s}^{m+1}, \vz^e)\\
\end{aligned}
\end{equation}
The contribution of a single sample $\vz_i^r=(\vx_i^r,y_i^r)$ in the training batch $\vz_r$ is defined as follows:
\begin{equation}
\small
\begin{aligned}
 \mathcal{I}(\vz_i^r,\vz^e)\approx (\eta_{s}\nabla_{\theta_{s}}\mathcal{L}_{s}(\theta_{s}^m , \theta_{t}^m, \vz_i^r))^\intercal\nabla_{\theta_{s}}  \mathcal{L}_{\text{val}}(\vz^{e},\theta_{s}^{m+1})
 \label{eqn:influence}
\end{aligned}
\end{equation}

By excluding loss irrelevant to the teacher in \cref{eqn:influence}, we define the distillation influence of $\vz^r_i$ to be:
\begin{equation}
\small
\begin{aligned}
 \mathcal{I}_{\text{distill}}(\vz^r_i,\vz^e) =&
 \nabla_{\theta_{s}}\mathcal{L}_{\text{ce}}(T(\vx^r_i;\theta_{t}^m),S(\vx^r_i;\theta_{s}^m))^\intercal \\ 
 &\nabla_{\theta_{s}}\mathcal{L}_{\text{ce}}(\vy^{e}, S (\vx^{e}; \theta_s^{m+1}))
\end{aligned}
\end{equation}

\section{Approximation Methods}
\label{sec:approximation}

Here, we efficiently approximate this gradient similarity using a Taylor expansion:
\begin{equation}
\small
\begin{aligned}
&\nabla_{\theta_{t}} \frac{1}{B^r} \sum_{i=1}^{B^{r}}w_i L_{\text{ce}} (T(\vz_i^r,\theta_{t}),S(\vz_i^r,\theta_{s})) \\
&=  \frac{1}{B^r} \sum_{i=1}^{B^r}
\nabla_{\theta_{t}}L_{\text{ce}} (T(\vx_i^r;\theta_{t}^m),S(\vx_i^r;\theta_{s}^m)) \\
&\nabla_{\theta_{s}}\mathcal{L}_{\text{ce}}(\vy^e, S (\vx^e; \theta_s^{m+1}))^\intercal \\
&\nabla_{\theta_{s}}\mathcal{L}_{\text{ce}}(T(\vx_i^r;\theta_{t}^m),S(\vx_i^r;\theta_{s}^m)) \\
&\approx \frac{1}{B^r} \sum_{i=1}^{B^r}
\nabla^2_{\theta_{s},\theta_{t}}{\mathcal{L}_{\text{ce}}( T(\vx_i^r;\theta_{t}^m),S(\vx_i^r;\theta_{s}^m))}  \\
&\nabla_{\theta_{s}}\mathcal{L}_{\text{ce}}(\vy^{e}, S (\vx^{e}; \theta_s^{m+1})) \\ 
&\approx \nabla_{\theta_{t}}\frac{1}{B^r} \sum_{i=1}^{B^r} \bigg[\frac{ \mathcal{L}_{\text{ce}}(T(\vx_i^r;\theta_{t}^m),S(\vx_i^r;\theta_{s}^+))}{2 \epsilon} - \\
&\frac{\mathcal{L}_{\text{ce}}(T(\vx_i^r;\theta_{t}^m),S(\vx_i^r;\theta_{s}^-))}{2\epsilon}\bigg]
\end{aligned}
\end{equation}
where $\theta_{s}^\pm = \theta_{s} \pm \epsilon \nabla \mathcal{L}_{\text{ce}}(\vy^e, S (\vx^e; \theta_s^{m+1}))$ and $\epsilon$ is a small scalar.

\section{A Closer Look at Meta Distillation}
\label{sec:meta_distillation}
In meta distillation, the loss on the validation set with respect to the teacher can be derived as follows:

\begin{equation}
\resizebox{\linewidth}{!}{
$\begin{aligned}
& \nabla_{\theta_{t}} \mathcal{L}_{\text{ce}}(\vy^{e}, S (\vx^e
; \theta_s^{m+1})) \\
&\quad= \nabla_{\theta_{t}}\mathcal{L}_{\text{ce}}(\vy^{e},  S (\vx^{e}; \theta_{s}^{m} - \eta_{s} \nabla_{\theta_{s}}\mathcal{L}_{\text{s}}(\theta_{s}^m,\theta_{t}^m,\vz^r))) \\
&\quad= \nabla_{\theta_{t}} (\theta_{s}^{m} - \eta_{s} \nabla_{\theta_{s}}\mathcal{L}_{\text{s}}(\theta_{s}^m,\theta_{t}^m,\vz^r)) \nabla_{\theta_{s}}\mathcal{L}_{\text{ce}}(\vy^{e}, S (\vx^{e}; \theta_s^{m+1})) \\
&\quad= \nabla_{\theta_{t}} ( - \eta_{s} \nabla_{\theta_{s}}\mathcal{L}_{\text{s}}(\theta_{s}^m,\theta_{t}^m,\vz^r))\nabla_{\theta_{s}}\mathcal{L}_{\text{ce}}(\vy^{e}, S (\vx^{e}; \theta_s^{m+1})) \\
&\quad= \nabla_{\theta_{t}} ( - \eta_{s}(1-\alpha) \nabla_{\theta_{s}}\mathcal{L}_{\text{ce}}(T(\vx^r;\theta_{t}^m),S(\vx^r;\theta_{s}^m))) \\
&\quad\qquad\nabla_{\theta_{s}}\mathcal{L}_{\text{ce}}(\vy^{e}, S (\vx^{e}; \theta_s^{m+1})) \\
&\quad= -\eta_{s}(1-\alpha) \nabla^2_{\theta_{s},\theta_{t}}{\mathcal{L}_{\text{ce}}( T(\vx^r;\theta_{t}^m),S(\vx^r;\theta_{s}^m))} \\
&\quad\qquad\nabla_{\theta_{s}}\mathcal{L}_{\text{ce}}(\vy^{e}, S (\vx^{e}; \theta_s^{m+1}))\\
&\quad\approx -\eta_{s}(1-\alpha)\nabla_{\theta_{t}} {\mathcal{L}_{\text{ce}} (T(\vx^r;\theta_{t}^m),S(\vx^r;\theta_{s}^m))}  \\
&\quad\qquad\nabla_{\theta_{s}}\mathcal{L}_{\text{ce}}(T(\vx^r;\theta_{t}^m),S(\vx^r;\theta_{s}^m))^\intercal  \\
&\quad\qquad\nabla_{\theta_{s}}\mathcal{L}_{\text{ce}}((\vy^{e}, S (\vx^{e}; \theta_s^{m+1}))) \\
&\quad\approx -\eta_{s} (1-\alpha)h\nabla_{\theta_{t}} {\mathcal{L}_{\text{ce}} (T(\vx^r;\theta_{t}^m),S(\vx^r;\theta_{s}^m))},
\end{aligned}$}
\end{equation}
where
\begin{equation*}
\small
\begin{aligned}
 h=&\nabla_{\theta_{s}}\mathcal{L}_{\text{ce}}(T(\vx^r;\theta_{t}^m),S(\vx^r;\theta_{s}^m))^\intercal \\
 &\nabla_{\theta_{s}}\mathcal{L}_{\text{ce}}(\vy^{e}, S (\vx^{e}; \theta_s^{m+1})).
\end{aligned}
\end{equation*}
\section{Hyperparameters}
\label{sec:hyperparameters}
\begin{table}[H]
\begin{tabular}{@{}lc@{}}
\toprule
{Hyperparameter} &   \\
\midrule
$\alpha$ & $0.6$ \\
maximum sequence length & $128$ \\
distillation temperature & $1$ \\
fine-tuning epochs & $6$ \\
student learning rate & $1e-4, 3e-5, 5e-5$ \\
batch size & $32$ \\ \bottomrule
\end{tabular}
\caption{Hyperparemeters in the experiments.}
\label{tab:hyperpara}
\end{table}

For our method, online distillation and meta distillation baselines, we fix the teacher learning rate at $3e-5$.

\section{More ablation study}
\label{sec:more_ablations}
\subsection{Datasets for Student's Feedback}
In our method, we utilize the feedback from the student model on the provided validation set of GLUE datasets directly. In this section, we investigate the impact of utilizing feedback derived from a new validation set that has been separated from the original training set.

We random sample 5 \% and 10 \% samples of the training set to generate a new validation set respectively. Then we apply our method to the new training set.

\begin{table}[H]
    \centering
    \resizebox{\linewidth}{!}{
    \begin{tabular}{lcccccccc}
    \toprule
    Ratio & MRPC & RTE & SST-2  & MNLI  & QNLI  & QQP & \\
    & F1/Acc. & Acc. & Acc.  & Acc. & Acc. & F1/Acc. & \\
\midrule
5 \% & 86.9/81.9 & 65.8 & 91.8 & 83.3/82.4 & 90.0 & 71.3/88.9 \\
10 \% & 86.7/81.0 & 64.5 & 92.4 & 83.1/82.2 & 89.8 & 71.0/89.0 \\
\bottomrule
\end{tabular}}
\caption{Experimental results on the test set of GLUE in the setting of teacher's utilizing feedback derived from a new validation set split from the training set. 5 \% and 10 \% indicates the proportion of the number of samples in the new validation set to the original training set.}
\label{tab:fb-abl}
\end{table}

The data used to measure the generalization of the student, whether it be from an existing validation set or a newly separated set, remains informative in both cases. As such, it is reasonable to expect that the feedback provided by the student to the teacher would not exhibit significant differences between the two sources.

Our experiments demonstrate that utilizing feedback from a validation set, whether pre-existing or newly separated from the training set, does not lead to significant variations in performance. However, it should be noted that the number of training samples may play a role in the results. When a subset of the training set is selected to form a new validation set, the number of training samples is reduced. This reduction may lead to overfitting in datasets of small or medium size, as there is not enough data information provided to the model. Conversely, in large datasets, the number of samples is sufficient to encompass a substantial portion of the data information, thus having minimal impact on the results.

\begin{figure}[t]
    \centering
    \includegraphics[width=0.9\linewidth]{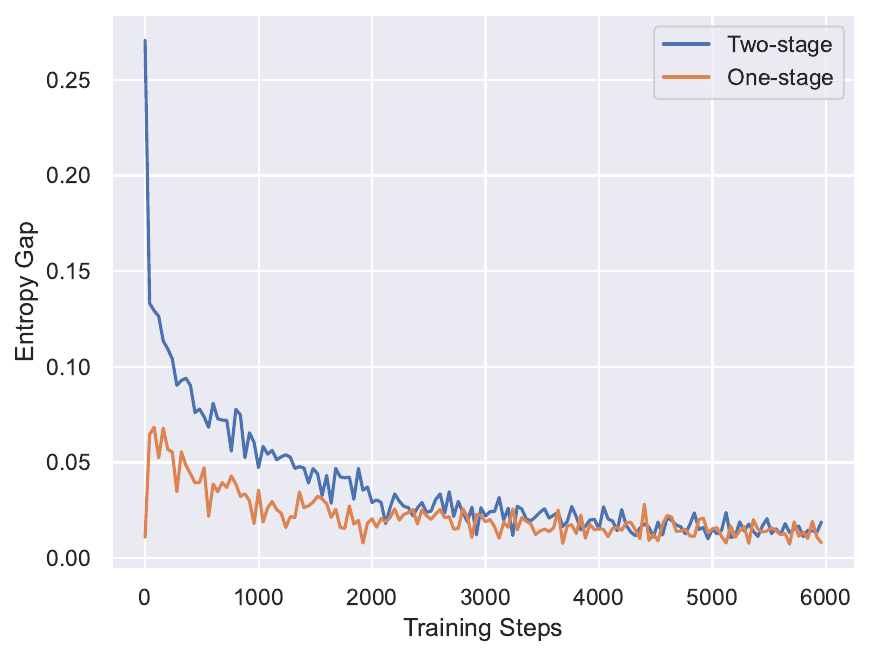}
    \caption{The entropy gap between the teacher and the student on the SST-2 training set for two-stage and one-stage training strategies. We only keep the loss with respect to ground truth labels in \cref{eqn:teacher} to train the teacher. We follow ~\citep{shi2020learning} to initialize both the teacher and student's classifier as zero in the one-stage setting.}
    \label{fig:exp_teacher-state}
\end{figure}

\begin{figure}[t]
    \centering
    \includegraphics[width=1\linewidth]{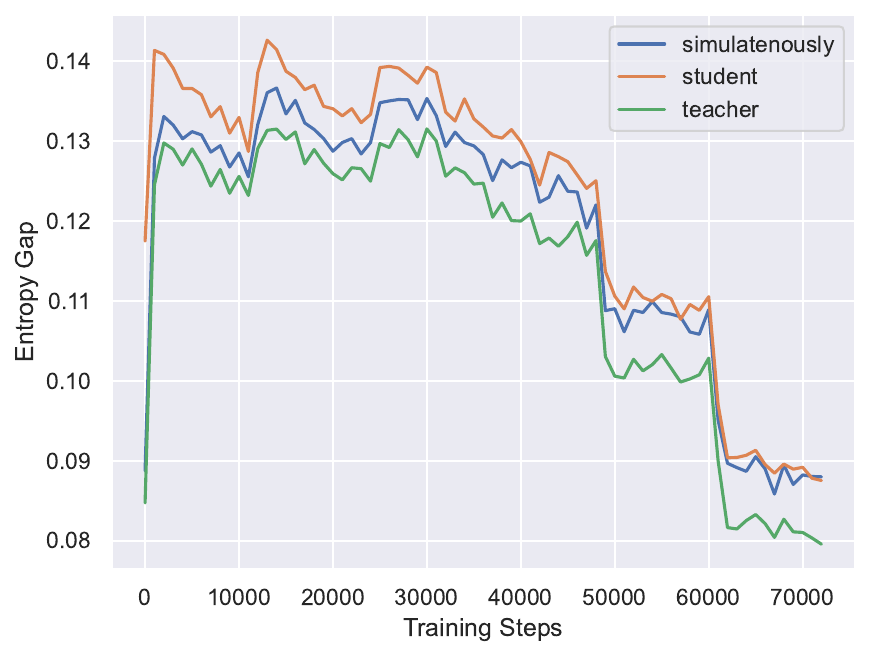}
    \caption{A comparison of the entropy gap on the MNLI training set with different orders of updating the teacher.`teacher' denotes updating the teacher model followed by the student model. `student' is the opposite. And `simulatenously' denotes updating the teacher and the student simulatenously.}
    \label{fig:exp_update_choice}
\end{figure}
\subsection{Ratio of Teacher's Self-evolution}
A student-friendly teacher should strike a balance between self-evolution and knowledge transfer. It is believed that an excessive focus on self-evolution may result in neglect of feedback provided by the student, leading to instruction that is not centered on the student's needs. Conversely, inadequate focus on self-evolution may prevent the teacher from improving their own abilities, resulting in suboptimal instruction for the student. In either scenario, the outcome is not conducive to fostering a student-friendly environment.

Therefore, we ablate on the ratio of the teacher's self-evolution to see how it contributes to the performance of the student. $\alpha$ is the ratio of the teacher's loss with respect to ground truth in \cref{eqn:t_loss}. We set it from \{1.0,0.8,0.6,0.4\}. 
\begin{table}[H]
    \tiny
    \centering
    \resizebox{\linewidth}{!}{
    \begin{tabular}{lcccccc}
    \toprule
    $\alpha$ & MRPC & RTE & SST-2  & MNLI   \\
    & F1/Acc. & Acc. & Acc.  & Acc.  \\
\midrule
1.0 & 87.0/81.9 & 66.1 & 92.3 & 83.0/82.1 \\
0.8 & 87.5/82.9 & 66.5 & 92.6 & 83.3/82.5 & \\
0.6 & \textbf{88.1/83.3} & \textbf{67.4} & \textbf{93.4} & \textbf{83.4/82.5} \\
0.4 & 87.5/82.8 & 66.1 & 92.2 & 83.3/82.5 & \\
\bottomrule
\end{tabular}}
\caption{Experimental results on the test set of four GLUE datasets. $\alpha$ is the ratio of teacher's self-evolution.}
\label{tab:alpha-abl}
\end{table}

In \cref{tab:alpha-abl},  the performance of the student exhibits a unimodal distribution, which is in agreement with our proposed assumption. Specifically, the results indicate that when the ratio of the teacher's self-evolution is set at 0.6, the performance of the student is optimal.

\section{Analysis}
We further discuss some design choices of current methods, including the initialization state of the teacher and the updating order of the teacher and student models. Following~\citep{guo2020reducing}, we apply the entropy gap to evaluate these design choices.

\subsection{Impact of the Teacher’s initial state}

While vanilla distillation and meta distillation employ a two-stage training approach, online distillation and LGTM employ a one-stage joint training strategy for the teacher and student models. The key difference is whether to involve fine-tuning the teacher network on target task. In this study, we investigate the impact of the teacher network's state on the student network.

A teacher network initialized in the same state as the student network can maintain the student network's progress at all times, but its capabilities may be relatively weak. In contrast, a converged teacher network has superior performance but also a larger gap, which can prevent the student network from gaining knowledge effectively. 

As show in \cref{fig:exp_teacher-state}, a lower initial confidence gap between the teacher model and the student model leads to more efficient knowledge transfer. When the initial ability gap is relatively high, it takes more iterations for the student model to catch up to the fine-tuned teacher model. In contrast, when the initial ability gap is lower, a teacher model initialized at the same state as the student model is able to transfer knowledge to the student more quickly. Specifically, in the early stages, the teacher model focuses more on self-evolution than knowledge transfer, causing the entropy gap to increase. Then, the teacher model shifts its focus towards knowledge transfer, resulting in an increasing and then decreasing trend in the entropy gap.

\subsection{Prioritizing the Teacher or Student}
Online distillation and meta distillation and LGTM all use bi-level optimization. However, online distillation and LGTM updates the teacher network followed by the student network, while meta distillation updates the student network followed by the teacher network. In this section, we study the optimal order for updating the teacher network and student network in knowledge distillation.

As shown in \cref{fig:exp_update_choice}, updating the teacher model first could lead to a 
lower entropy gap and faster convergence speed. We assume that the teacher could formulate an appropriate `teaching plan' for the student in this updating order. 

The teacher should strive to guide the student to identify the most important samples and information, to help the student develop a deep and general understanding of the task. Furthermore, the teacher should also take into consideration that some samples may be difficult for the teacher itself to classify or understand. And for those samples, a lower criterion should be set for the student, which may form a more student-friendly decision boundary. 

Therefore, the teacher's output serves as a dynamic learning target for each sample. By updating based on the student's feedback in advance, the teacher is able to reach a state that is optimal for the student's learning. In this case, the teacher could provide an appropriate learning signal. Leveraging this updated supervision signal, the student could make up for the ability gap faster. For the other two updating orders, the teacher hasn't updated yet, lacking of making trade-offs between the samples that are more beneficial for generalization and those that are more challenging to learn from. This may lead to a certain degree of lag in knowledge transfer, resulting in a larger entropy gap between the student and the teacher.

\end{document}